\documentclass[10pt,twocolumn]{article}

\usepackage[a4paper,margin=0.8in]{geometry}
\usepackage{times}
\usepackage{flushend}
\usepackage{graphicx,color}
\usepackage{caption}
\usepackage{url}
\usepackage{hyperref}
\usepackage[numbers,sort&compress]{natbib}
\usepackage{titlesec}
\usepackage{setspace}
\usepackage{ulem}
\usepackage{fancyhdr}

\titleformat{\section}{\bfseries\large}{\thesection.}{0.5em}{}
\titleformat{\subsection}{\bfseries\normalsize}{\thesubsection.}{0.5em}{}
\title{\vspace{-1em}\textbf{Beyond the Failures: \\Rethinking Foundation Models in Pathology}}
\author{Hamid R. Tizhoosh\\
\small Kimia Lab, Department of Artificial Intelligence and Informatics, Mayo Clinic, Rochester, MN, USA\\ Email: \emph{tizhoosh.hamid@mayo.edu}}
\date{}

\begin{document}
\maketitle
\begin{abstract}
\noindent
Despite their successes in vision and language, foundation models have stumbled in pathology, revealing low accuracy, instability, and heavy computational demands. These shortcomings stem not from tuning problems but from deeper conceptual mismatches: dense embeddings cannot represent the combinatorial richness of tissue, and current architectures inherit flaws in self-supervision, patch design, and noise-fragile pretraining. Biological complexity and limited domain innovation further widen the gap. The evidence is clear—pathology requires models explicitly designed for biological images rather than adaptations of large-scale natural-image methods whose assumptions do not hold for tissue.
\end{abstract}


\section{Introduction}
Foundation models (FMs) have transformed image and text processing by leveraging massive data corpora and self-supervised learning \cite{bommasani2021opportunities, awais2025foundation}. Their success in general computer vision inspired enthusiasm for analogous applications in histopathology, where the potential for automating diagnostic reasoning and multimodal retrieval is expected to be immense \cite{bilal2025foundation,xiong2025survey}. Indeed, pathology seemed an ideal candidate for FM-driven progress: vast archives of digitized whole-slide images, rich morphological variation, and deep links to genomic and clinical context. \textcolor{black}{A good FM in pathology should satisfy some criteria including  cross-center robustness beyond site-specific artifacts; stability under clinically plausible perturbations (rotation, stain, scanner); calibrated confidence and error consistency; meaningful slide-level retrieval; and feasibility of specialization with moderate labeled data.}
However, the emerging evidence is unambiguous: pathology FMs are not delivering on their promise. Under realistic clinical evaluation, they consistently reveal structural weaknesses that cannot be dismissed as temporary growing pains. 
Across essentially every benchmark evaluation, pathology FMs reveal the same pattern: only modest diagnostic accuracy, fragile cross-institutional generalization, and an unsettling sensitivity to trivial changes in rotation, color, or small adversarial perturbations — weaknesses that would be unacceptable in any clinical setting.
More recent works further clarify why these models struggle: dense single-vector embeddings—the dominant representation used in pathology FMs—face provable theoretical limits, unable to encode the combinatorial complexity of tissue morphology even under ideal training conditions.
In parallel, new findings on catastrophic inheritance show that even small amounts of label noise introduced during pretraining can irreversibly collapse the statistical geometry of the learned representation space, degrading downstream performance in medical tasks despite appearing effective on upstream objectives. 
Taken as a whole, the evidence shows that pathology FMs are not hindered by superficial problems—they are limited by profound conceptual mismatches in representation, learning, and cross-domain transfer that current architectures simply cannot overcome.

This short comment analyzes these limitations and investigates the biological, mathematical, and architectural factors that contribute to their shortcomings, with the goal of informing future improvements in model design.

\section{Empirical Evidence of Weaknesses}
While not exhaustive, the following summaries highlight several recent studies that have reported weaknesses in foundation models to analyze tissue images in histopathology.

\textbf{Low Accuracy and Inconsistent Clinical Performance -- }
Alfasly \textit{et al.}~\cite{alfasly2025low} evaluated several leading pathology foundation models—including UNI, GigaPath, and Virchow—on 11,444 whole-slide images from 23 organs and 117 cancer subtypes in TCGA using a zero-shot retrieval framework based on Yottixel’s patch-embedding approach. 
Despite their size and ambition, these models managed only 40–42\% macro-averaged F1 for top-5 retrieval and showed wide, clinically unacceptable variability from one organ system to another: kidneys reached up to ~68\% (top-1 F1)\footnote{Outlier organ performance, such as the unusually high kidney F1 scores, may partly reflect genuine morphological distinctiveness—but could also indicate \textbf{data leakage} or hidden technical bias.
In pathology FMs, these effects are common and often hard to disentangle, so all high outlier results warrant careful patient-level and site-level validation.}, whereas lungs dropped to $\approx$21\%. Aggregating patches into single WSI-level embeddings did not improve results and sometimes degraded them, suggesting loss of spatial information. 
Overall, even though pathology FMs outperform CNN baselines, their accuracy remains far too low for clinical relevance, and their poor cross-tissue generalization suggests that they are still modeling texture noise instead of the morphological signatures that pathologists actually use to diagnose disease. \textcolor{black}{One must distinguish relative gains over prior baselines from absolute clinical adequacy. The analysis here is grounded in zero-shot or weakly supervised evaluation, cross-institutional generalization, robustness to clinically plausible perturbations, and retrieval-based performance, rather than heavily optimized downstream pipelines. Strong results on curated benchmarks or coarse labels may overestimate clinical readiness by masking representational fragility.}

\textbf{Lack of Robustness and Site Bias -- }
De Jong \textit{et al.}~\cite{dejong2025robustness} systematically evaluated the robustness and generalization of ten leading pathology foundation models across multiple institutions and datasets using a newly defined \emph{Robustness Index} (RI), which quantifies whether model embeddings cluster more strongly by biological class or by medical center. The RI compares within-class versus within-center similarity (with RI$>$1 indicating true biological robustness). Among all tested models, only Virchow2 achieved RI $\approx$ 1.2—meaning biological structure dominated site-specific bias—whereas all others had RI $<$ 1 (e.g., UNI $\approx$ 0.9, Phikon-v2 $\approx$ 0.7). Embeddings from most models, therefore, grouped primarily by hospital or scanner rather than by cancer type, leading to large performance drops on unseen centers. 
The study ultimately shows that today’s pathology FMs are fragile systems that are readily confounded and lack the domain robustness required for medicine—making cross-institutional validation and bias-resistant architectures non-negotiable before these models can be taken seriously for clinical use.

\textbf{Geometric Fragility -- }
Elphick \textit{et al.}~\cite{elphick2024rotation} investigated 12 self-supervised pathology foundation models and assessed how well their latent representations remain stable when image patches are rotated. The authors apply rotations in 15° increments from 0° to 360° on patches extracted from the TCGA-KIRC dataset and compute two metrics to quantify invariance: mean mutual k-nearest neighbours (m-kNN) and mean cosine distance between embeddings of non-rotated versus rotated patches. They report that PathDino (a small model with fewer than 10M parameters \cite{alfasly2024rotation}) achieved the highest m-kNN score of 0.85, making it the most rotation-invariant by that measure, and that Hibou‑L achieved the lowest cosine distance of 0.016, indicating the best alignment for that metric. In contrast, UNI achieved an m-kNN of almost 0.73, whereas Virchow produced the lowest score ($\approx$0.53). Importantly, the results show statistically that models trained with explicit rotation augmentation significantly outperform those without (t = 6.91; $p < 0.0001$ for m-kNN; $t = -8.88$; $p < 0.0001$ for cosine distance). 

The finding that a lightweight model—roughly thirty times smaller than the smallest pathology foundation model—exhibits far superior rotation robustness should serve as a wake-up call. It exposes a fundamental flaw in the current FM paradigm: scale alone does not guarantee clinical reliability, and may, in fact, conceal brittle, ill-structured representations that fail under basic geometric transformations.

\textbf{Resource Burden and Fragile Adaptation -- }
Mulliqi \textit{et al.}~\cite{mulliqi2025limitations} conducted a large-scale study using over 100,000 prostate biopsy slides (from 7,342 patients across 15 sites in 11 countries) to compare two pathology foundation models (FMs) against a task-specific (TS) end-to-end model for prostate cancer diagnosis and Gleason grading. They found that although the FMs offered utility in data-scarce settings, when enough labeled data were available, the TS model matched or even outperformed the FMs. \textcolor{black}{Critically, the foundation models consumed up to 35× more energy than the task-specific model, raising sustainability concerns when considered within a realistic clinical lifecycle. This comparison is not intended to equate one-time foundation pretraining with a single downstream training run, but rather to highlight the marginal energy cost of achieving comparable clinical performance in a validated setting. Energy use must be considered in terms of deployment-relevant factors, including the need for repeated fine-tuning or adaptation across tumor types and institutions, inference-time footprint, and the limited amortization benefit when foundation models do not substantially outperform smaller task-specific alternatives. Hence, computational scale alone may not guarantee sustainable or clinically efficient AI in pathology.}

Consuming orders of magnitude more memory while failing to surpass smaller task-specific models, current FMs undermine—rather than enable—the promise of democratizing AI into accessible and clinically reliable technologies.

\textbf{Linear Probing: Because Fine-Tuning Does Not Work
 -- }
Although foundation models are often promoted for their flexibility and ``emergent'' adaptability to new tasks, in computational pathology their downstream use is overwhelmingly limited to linear probing—training a shallow linear classifier on frozen embeddings rather than fine-tuning the model itself. This dependency arises because most pathology FMs are too large, memory-intensive, and unstable to fine-tune on moderate-sized datasets typical of clinical research (often hundreds to a few thousand slides). Recent studies confirm that full fine-tuning frequently degrades accuracy relative to linear probing due to overfitting and catastrophic forgetting \cite{mulliqi2025limitations,liang2025benchmark}. Yet this pragmatic retreat stands in stark contrast to the foundational premise of the FM paradigm, as articulated in Bommasani et al.'s seminal paper \cite{bommasani2021opportunities} and many subsequent multimodal works that promise large pretrained systems should enable \emph{zero-shot} and easily fine-tuned adaptation across domains. In pathology, the FM promise falls apart: most models operate only as frozen feature banks that need to be linearly probed, a scenario akin to \textbf{purchasing a Ferrari that does not run and then relying on a bicycle to tow it}. Far from being foundational, these models expose a striking mismatch between the narrative of universality and the realities of clinical AI deployment. \textcolor{black}{FMs are not inherently untunable, but that instability, compute cost, and risk of degradation often lead practitioners to rely on linear probing in practice. Parameter-efficient fine-tuning methods (e.g., LoRA \cite{hu2022lora}) may mitigate memory and stability issues, but they do not fully resolve domain mismatch or inherited bias.}

\textbf{Security and Safety Vulnerabilities -- }
Wang \textit{et al.}~\cite{wang2025utap} introduced \textit{Universal and Transferable Adversarial Perturbations (UTAP)}, imperceptible noise patterns that collapse FM embeddings across architectures. These universal attacks threaten clinical reliability. The paper shows that visually imperceptible perturbations at modest pixel bounds can severely degrade representation quality, collapsing accuracy from $\approx\! 97$\% to $\approx 12$\% on the attacked model and transferring across black-box models (e.g., $96.42\%\rightarrow48.69\%$, $97.23\%\rightarrow25.52\%$), while also harming out-of-distribution datasets. These attacks break the learned feature manifold rather than merely flipping labels, are easy to learn ($\approx900$ patches, a few minutes on an RTX,4090), and generalize widely. Thus, the high clean-test accuracy of ViT-based patch encoders can be deceptive, leaving deployed pathology systems vulnerable and posing significant safety risks to patients.

In histopathology, these perturbations pose significant safety risks and  have a dual interpretation that goes beyond malicious attacks that are clearly a security risk. They have a real-world noise analogue; they approximate the small, systematic variations that arise naturally in the imaging pipeline: 

$\circ$ Differences in H\&E staining

$\circ$ Scanner optics and illumination variability

$\circ$  Compression artifacts

$\circ$ color normalization and rescaling

$\circ$ Slide preparation imperfections\\ ----- \emph{dust, bubbles, section thickness, etc.}

$\circ$  Downstream digital processing \\---- \emph{gamma correction, color-space conversion} 

In this view, adversarial perturbations are not malicious but \textbf{diagnostic stress tests} that reveal how sensitive a model is to minor pixel-level changes that can—and do—occur in laboratory and acquisition workflows. \textcolor{black}{Consistent with this perspective, recent studies \cite{henriksen2026enabling} show that histopathology foundation models are highly sensitive to non-biological variations such as scanner and preparation differences, which can dominate their representations and lead to inconsistent predictions.} 

That pathology FMs remain vulnerable to perturbations as trivial as slight color shifts should give the field pause: a model so easily destabilized by routine laboratory and imaging variation cannot be trusted to safeguard patients, and its deployment in clinical workflows would be a deeply unsettling proposition.

\section{Etiology of the FM Limitations and the Path Forward}
This section offers a conceptual and holistic analysis of why foundation models in pathology underperform, probing the cognitive, methodological, and epistemic assumptions embedded in current AI paradigms. Where possible, it also points toward a path forward, grounded in available empirical evidence, historical experience, and the best-informed projections for clinical AI.

\begin{figure}
    \centering
    \includegraphics[width=0.99\linewidth]{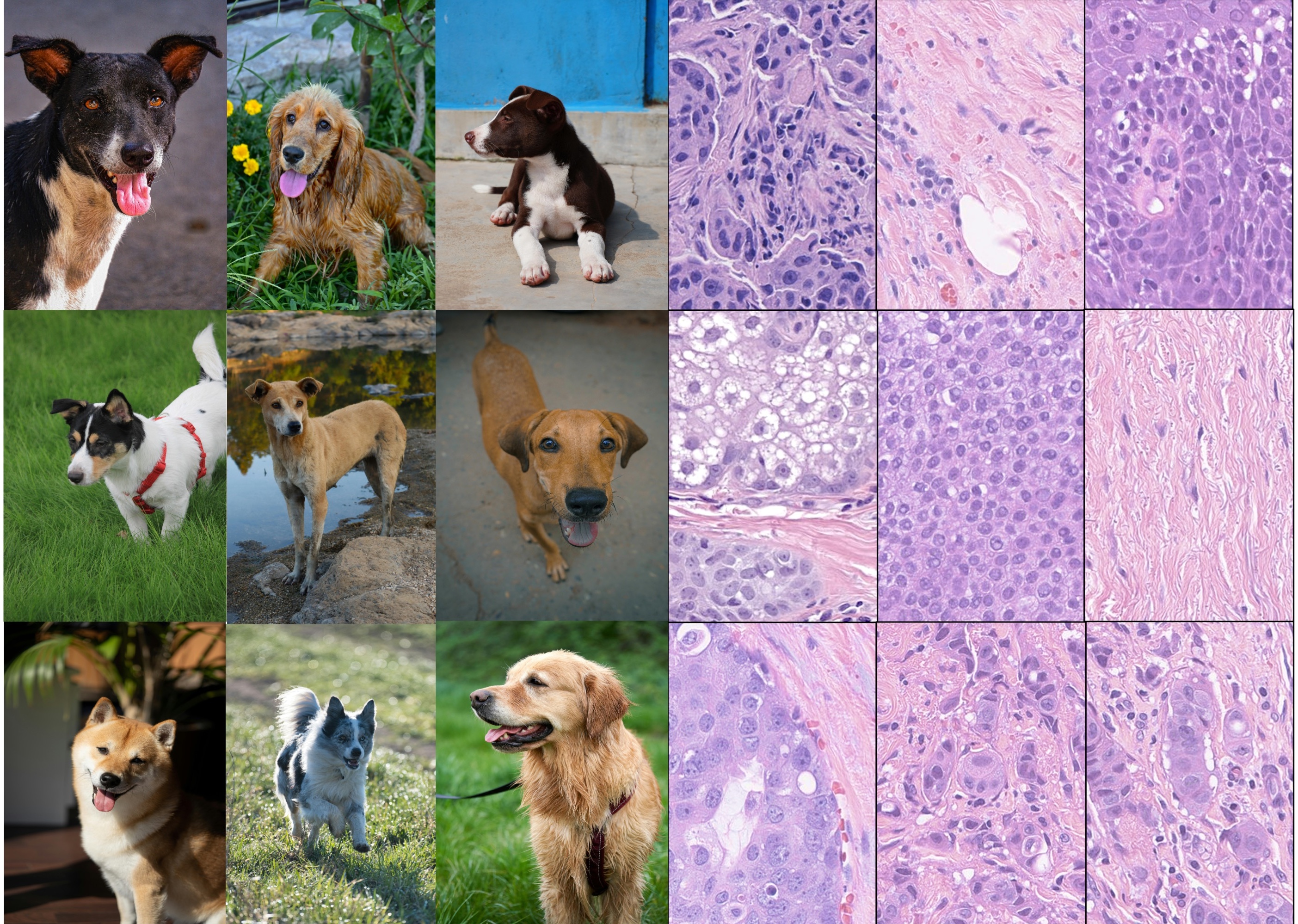}
    \caption{AI models can recognize dogs and even distinguish among breeds—tasks that children can perform with ease. In contrast, recognizing complex tissue patterns in pathology requires an adult with more than a decade of specialized education and training.} 
    \label{fig:dogtissue}
\end{figure}

\textbf{Underestimating the Complexity of Human Tissue -- }
The AI community often underestimates the semantic complexity of tissue morphology (Fig.\ref{fig:dogtissue}). A child learns to recognize dogs by age two and breeds by seven (a task at which AI performs exceptionally well)  \cite{eretova2020can}; A pathologist—a highly trained human adult—typically requires more than twelve years of education to distinguish cancer subtypes based on tissue morphology. Unlike natural images, tissue interpretation depends on context, scale, and clinical correlation—far beyond simple object recognition.
The \underline{road ahead} must couple the strengths of deep texture analysis with purposeful integration of histology-specific knowledge—an essential step if future models are to move beyond surface patterns and capture true diagnostic morphology. 

\begin{figure}
    \centering
    \includegraphics[width=0.99\linewidth]{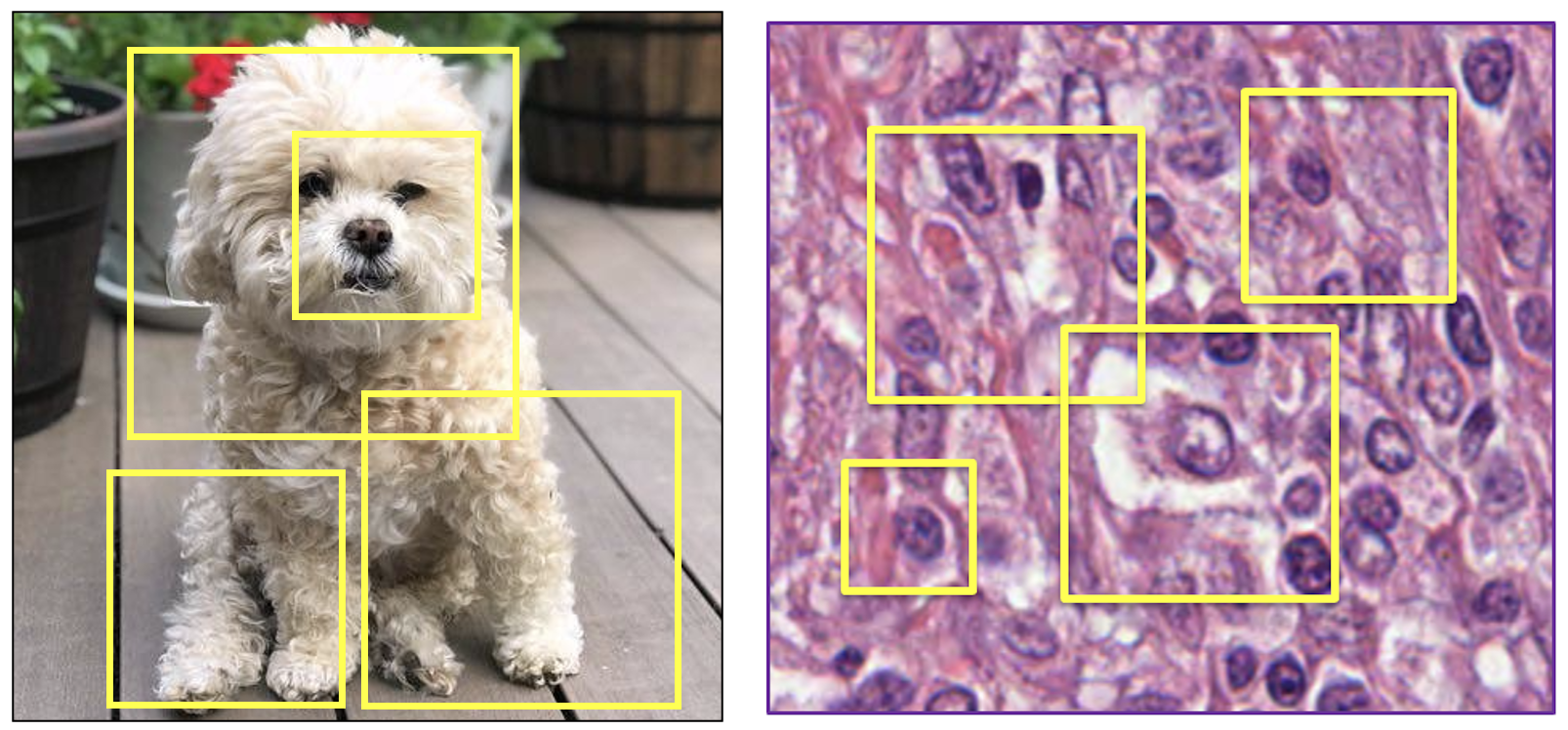}
    \caption{Self-supervised learning often rests on the implicit assumption that each image represents a single, coherent object—an assumption that fails in histopathology, where multiple heterogeneous tissue structures coexist within the same field of view.}
    \label{fig:sslcrops}
\end{figure}

\textbf{Ineffective Self-Supervision for Tissue Images -- }
\textcolor{black}{When we focus on learning objectives and loss functions we observe that} most early self-supervised learning frameworks were developed and validated on single-object datasets such as ImageNet, making them less suited for complex, structure-rich images like histopathology slides, which lack discrete, well-defined objects (Fig.\ref{fig:sslcrops}). The ``local-global crop'' assumption fails in tissue slides, where patches contain mixed or irrelevant content. As a result, models learn stain texture instead of biological patterns, weakening generalization.
The \underline{path forward} is not to abandon self-supervision, but to tame it—through hybridization with complementary approaches and a renewed examination of how supervised and unsupervised topologies can be combined to reflect the realities of histological data. \textcolor{black}{This addresses failures of generic self-supervision—including weak semantic alignment, poor cross-institution generalization, label ambiguity, and instability during fine-tuning—by anchoring representation learning to clinically meaningful diagnostic concepts expressed by expert pathologists.}

\textbf{The Myth of the Universal Model -- }
According to the \emph{No Free Lunch theorem}, no single model excels across all problems. Expecting one FM to generalize to all organs and cancers ignores pathology’s heterogeneity. Benchmarks reveal wide organ-dependent performance swings, underscoring the limits of universal architectures. The \underline{path forward} is to reconsider our assumptions about foundation models in medicine and acknowledge that some degree of specialization during training may be necessary to align AI models more tightly with histopathology—without sacrificing organ- or subtype-level generalization.

\textbf{Architectural Excess and Occam’s Razor -- }
Modern foundation models (FMs) often pursue ever-greater depth and parameter scaling without demonstrable performance benefits. In line with Occam’s Razor, progress in pathology AI may instead depend on leaner, domain-structured architectures that embody the hierarchical and contextual organization of biological tissues, thereby enhancing both interpretability and generalization. \underline{Moving forward}, we must re-internalize a simple truth: bigger is not always better.

\textbf{Lack of Domain-Specific Innovation -- }
\textcolor{black}{When we focus on architectural choices, input representations, and inductive biases, } many pathology foundation models (FMs) are direct adaptations of general-purpose frameworks such as CLIP, DINO, or MAE, retrained or fine-tuned with pathology data. However, few incorporate domain-specific mechanisms such as magnification awareness, stain-invariant representations, or morphology-aware pretext tasks, resulting in methodological stagnation and limited engagement with the underlying biological domain \cite{li2025survey}. Despite the unique multi-scale and heterogeneous nature of tissue images, no major advances have been introduced in model topology, input preparation, or loss formulation to explicitly tailor foundation models to the structural and semantic characteristics of histopathology. The \underline{future} will arrive sooner once we stop following mainstream AI trends and start inventing specifically for medicine.

\textbf{Data Deficit and Scaling Limits -- }
CLIP was trained on 400 million image–text pairs; no pathology dataset approaches that scale. Even the largest multi-institutional archives offer fewer than one million WSIs, fragmented and inconsistently labeled. \textcolor{black}{Current foundation model scaling is driven primarily by patch count (several hundred or even thousands per slide), not by the number of WSIs. As such, even a modest cohort—on the order of 100k WSIs—may correspond to 100–200 million patches. However, given the limitations of patch-based processing and the small effective field of view, this scale must be put into perspective. Meaningful progress may still require substantially larger numbers of WSIs, not merely more patches. Furthermore, } privacy constraints exacerbate scarcity, capping the value of foundation-scale pretraining. The \underline{direction ahead} depends less on algorithms and more on leadership—from hospitals, governments, and cloud providers. Will nations invest in shared medical imaging repositories? Will hospitals reclaim autonomy through self-governed data centers? And will cloud providers finally make storage and compute accessible rather than extractive? These decisions will define whether reliable medical AI becomes a democratized public good or an elitist tool available only to well-resourced institutions

\begin{figure}
    \centering
    \includegraphics[width=0.99\linewidth]{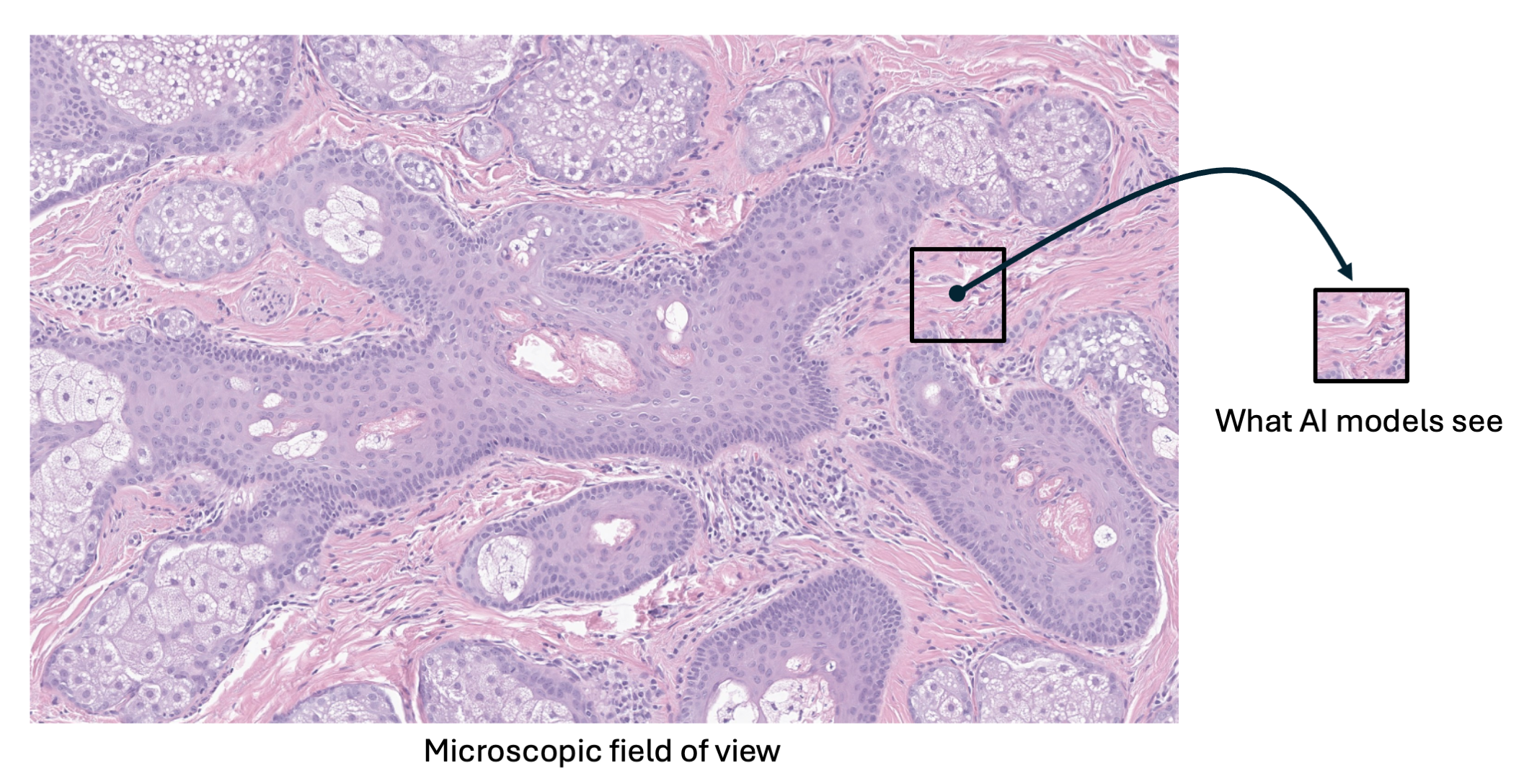}
    \caption{The field of view in light microscopy is traditionally quite large—approximately 2000 × 1500 pixels—and becomes vastly larger in whole-slide images (WSIs). In contrast, most AI models operate on small image patches, typically around 224 × 224 pixels.}
    \label{fig:vit16}
\end{figure}

\textbf{Patch Size and Field-of-View Mismatch -- }
A critical but widely overlooked issue is the patch-size mismatch between ViT architectures and diagnostic field of view. Most FMs use 224$\times$224-pixel patches, a convention inherited from ImageNet. Yet even low-end educational microscopes produce 2048$\times$1536-pixel (3 MP) views sufficient for diagnostic teaching (Fig. \ref{fig:vit16}). 
Such small tiles capture fine-grained micro-texture but fail to represent mesoscale tissue architecture, glandular context, or stromal organization. This design choice prioritizes computational convenience—perhaps even favoring rapid experimentation and publication—over biological realism. Consequently, many foundation models end up encoding superficial texture statistics rather than diagnostically meaningful morphology. To bridge this gap, adaptive or hierarchical patching strategies combined with multi-scale attention mechanisms are urgently needed to model both local patterns and global structural context. Vision Transformers (ViTs) split images into small patches (e.g., 16×16) to convert them into manageable token sequences for self-attention. This is computationally efficient but semantically costly: the model initially loses the global spatial structure (paramount for tissue morphology) and must learn it back from data. In pathology, where diagnostic meaning resides in multimagnification architecture, this design leads to \textbf{models that see textures but not tissues}. 

And by reducing gigapixel slides to tiny patches, the field has backed itself into a corner where it must construct increasingly bloated \textbf{aggregation} frameworks to stitch together unstable and poorly justified patch embeddings into something resembling a whole-slide representation. \textcolor{black}{Whole-slide–level models \cite{hemati2023learning,ding2025multimodal} process many or all patches of a WSI and some may reduce reliance on patch aggregation; however, these approaches remain insufficiently validated across institutions and clinical tasks, and their robustness and adaptability have yet to be established by the broader pathology community.}
The \underline{next frontier} is clear: models must learn from larger and more representative images (and not tiny patches), a non-negotiable requirement for gigapixel pathology slides that should be self-evident at this point. \textcolor{black}{This addresses failures related to the loss of meso- and macro-architectural context, reliance on fragile patch aggregation, poor slide-level reasoning, and sensitivity to tumor sparsity, where diagnostically relevant regions occupy only a small fraction of the tissue.}

\textcolor{black}{There is a wide range of multiple instance learning (MIL) approaches beyond linear probing, including attention-based, hierarchical, and graph-based aggregation strategies that have demonstrated practical utility in computational pathology. These methods can effectively boost task-level performance by modeling complex relationships among patches and capturing higher-order context at the slide level. However, such downstream aggregation can also compensate for deficiencies in the upstream foundation model representation, masking limitations in feature quality and semantic alignment. As a result, strong downstream performance may obscure representational fragility and complicate rigorous assessment of clinical robustness, cross-institution generalization, and transferability of the foundation model itself.}

\textbf{Catastrophic Inheritance} 
Sun et al. \cite{sunpretraining} present the first systematic analysis of catastrophic inheritance in medical foundation models — a phenomenon where small amounts of label noise during pretraining cause severe degradation when models are transferred to clinical downstream tasks. The authors show that even 5\% synthetic label corruption introduced during pretraining (on ImageNet-1K or YFCC15M) leaves downstream models consistently less robust in Camelyon17, HAM10000, and NIH-ChestXray, despite the fact that in-distribution accuracy may remain deceptively high.

Crucially, the paper shows that the problem is structural, not merely statistical. As noise increases (5\%, 10\%, 20\%, 30\%), the skewness and kurtosis of feature embeddings and logits collapse. These higher-order statistics — normally high in fine-grained medical data — shrink toward zero, indicating that the learned representation space becomes flattened, symmetric, and less discriminative. The paper demonstrates this collapse mathematically (without requiring new architecture) and empirically through controlled experiments. For example, on Camelyon17, the skewness mean drops from 6.00 → 3.61, and kurtosis mean drops from 110.02 → 57.28 under 30

The authors then propose a lightweight fine-tuning strategy called SKD (Skewness-Kurtosis Distributional Regularization) — adding two small scalar penalties to the loss function, restoring asymmetry (skewness) and sharpness (kurtosis) to the representation. Importantly, this approach does not modify the foundation model backbone — which is realistic since models like PLIP and MedCLIP are often frozen checkpoints that clinicians cannot retrain. SKD works with both supervised and contrastive models (ResNet-50, CLIP, ViT) and improves downstream performance across vision and language tasks, often by 3–7\% over linear probing and outperforming spectral-based noise mitigation (NML).

Although primarily demonstrated on medical images, the authors explicitly note that the theory applies to any domain where pretraining data is noisy and downstream tasks require fine-grained discrimination. Thus, the results generalize beyond pathology to radiology, dermatology, biomedical NLP, and other clinical AI settings.
In essence, the noise in pretraining is inherited and cannot be fixed by standard transfer learning. This leads to geometry collapse (low skew/kurtosis). But it can be mitigated by restoring distributional structure. This insight reframes pretraining noise as a \textbf{representational failure mode}, not just a drop in accuracy, and offers a domain-aware remedy that is computationally cheap, architecture-agnostic, and immediately relevant for clinical AI. The \underline{future trajectory}—however uncomfortable—must move beyond the crutch of ``ImageNet pretraining.'' Computational pathology has reached a point where researchers must roll up their sleeves and train models from scratch.

\section{Is there an Underlying Cause?}
We discussed the complexity of human tissue as a possible cause for the failures of FMs. We mentioned that self-supervision may be an ineffective learning method despite its success in non-medical domains. We also considered that developing a universal model for histology may not be a practical approach—the models could become too large, violating Occam’s razor. In addition, we pointed out the lack of innovation, particularly in the topology and learning schemes for histopathology. Despite using millions of small patches (e.g., 224×224), we still do not have sufficient data to scale and train models like CLIP. Furthermore, we noted that patch sizes may be too small given the historical field of view in microscopy. Most recently, as an etiological reason,  catastrophic inheritance has been suggested as a possible cause for the failures of deep models. The question remains: \underline{Is there an even bigger reason for FM failures?}

Weller et al. \cite{weller2025theoretical} present a rigorous and foundational critique of dense embedding models—systems that encode queries and documents into fixed-size vectors and rely on similarity search (e.g., dot product) for retrieval. Although much of today’s LLM ecosystem (such as RAG systems) depends on these embeddings for retrieving relevant contexts, the paper shows that this entire paradigm faces irreducible theoretical constraints. Using tools from \textbf{communication complexity} and \textbf{sign-rank theory}, the authors prove that for any embedding dimension $d$, there exist retrieval tasks that are \uline{impossible to represent—even with perfect training, complete supervision, or unlimited compute}. In particular, the number of distinct relevance patterns that embeddings can realize scales only polynomially with $d$, while real-world retrieval demands represent combinatorially structured relationships that grow exponentially. This means that even in a hypothetical best-case scenario — where vectors are directly optimized to fit the test set (``free embeddings'') — embeddings of size 768 or 1024 cannot fully encode all the retrieval patterns required for many tasks. Their experiments confirm this, showing that even with \emph{idealized optimization}, the upper bound on retrievable documents is limited to only a few million ($\approx$4M at d = 1024, $\approx$250M at d = 4096), far below the combinatorial requirements of many retrieval settings.

Crucially, although most empirical examples in the paper involve LLM-based text retrieval, the authors explicitly emphasize that these theoretical constraints apply \emph{equally to all embedding-based retrieval systems}, regardless of modality — including \textbf{image, audio, video, biometrics, medical imaging, and pathology embeddings}. When a model reduces a rich object (such as a slide, image patch, radiograph, or clinical record) to a single d-dimensional vector, the same mathematical limits apply. \textcolor{black}{This can be viewed as a fundamental representational bottleneck as a potential representational limitation under certain conditions: the structure of dense embeddings simply cannot may not always be able to grow expressive capacity fast enough to match the combinatorial space of real-world relevance patterns, especially in domains with multi-factor reasoning, spatial complexity, or semantic heterogeneity.}

Improvements in training data or architecture cannot fix these limits, and \underline{future retrieval systems} must move beyond single-vector embeddings toward multi-vector, sparse, hybrid, or cross-encoder models—a message that should strongly resonate with fields like computational pathology, where tissue complexity defies simple fixed-dimensional encoding. \textcolor{black}{This addresses failures arising from single-vector compression, including limited expressivity for heterogeneous tissue patterns, degraded retrieval fidelity, sensitivity to confounders, and the inability to represent multiple coexisting diagnostic features within a single slide.}

\textcolor{black}{The single-vector embedding argument is intended as a boundary condition rather than a complete explanation for current foundation models' performance. Dense embeddings are demonstrably powerful in many vision and language systems; however, histopathology places particular stress on such representations due to extreme class imbalance, subtle and often diffuse morphological variation, hierarchical spatial organization, and the presence of clinically meaningful but rare patterns. Tissue morphology in pathology may not inherently or universally “harder” than natural images in all respects. Rather, performance limitations arise from a mismatch between representational assumptions and clinical tasks—most notably in retrieval, robustness to perturbation, and cross-domain transfer—where compressing heterogeneous tissue signals into a single fixed-length vector can have a disproportionate impact in computational pathology.}

\section{Conclusions}
Foundation models have undeniably transformed vision and language, and their entry into medicine is often portrayed as the next inevitable leap. Yet the biological complexities of human tissue—and the interpretive reasoning that pathology demands—cannot be solved by scale or by inheritance from off-the-shelf topologies. The growing body of evidence, combined with new theoretical limits on dense embeddings and the discovery of catastrophic inheritance from noisy pretraining, leaves little room for ambiguity: current pathology FMs are not merely immature—they are fundamentally misaligned with the biological and analytical realities of the domain.

These limitations are not detours on the road to progress; they are obvious warning signs that the road itself must change. Progress will not come from further scaling data or parameters, but from models that reflect the multi-scale, spatially structured, and clinically grounded nature of human tissue. Embedding architectures must move beyond compressing gigapixel morphology into fixed-length vectors. Transfer learning pipelines must preserve—rather than collapse—the statistical geometry that carries diagnostic meaning. And evaluation must shift from upstream benchmarks to metrics that reflect real clinical reliability, domain robustness, and interpretive truth.

True innovation will require redefining what a ``foundation'' in medical AI should be. We need systems that perceive tissue as pathologists do: in context, across scales, rooted in genuine biological variation, and validated within transparent, rigorous clinical frameworks. Seen this way, current pathology FMs are not the destination but the first draft—prototypes that reveal, with some harsh clarity, what the field must leave behind and what it must build next. With conceptual reorientation and domain-specific design, future foundation models can evolve from computational showcases into trusted devices of diagnosis and discovery, closing the widening gap between algorithmic ambition and medical insight.

\textcolor{black}{Finally, to make validation criteria actionable and falsifiable, future pathology foundation models should be evaluated using standardized, clinically grounded protocols. These include cross-institution generalization (e.g., leave-one-center-out validation), robustness under controlled perturbations such as rotation, stain variation, compression, and scanner shifts, and calibration metrics assessing confidence reliability (e.g., expected calibration error and prediction consistency). In addition, retrieval-based evaluation should be incorporated, using metrics such as Top-1 accuracy and majority agreement among Top-3 or Top-5 retrieved cases to assess clinical relevance. While precise numerical thresholds may vary by task and dataset, consistent application of these evaluation protocols provides a practical framework to determine whether a model meets minimal standards for clinical robustness, generalization, and interpretability.}

\section{Author Contributions}

Hamid Tizhoosh is the sole author.

\section{Acknowledgment}
The author would like to thank Dr. Saghir Alfasly and Dr. Wataru Uegami for valuable discussions.

\section{Funding}
No funding.

\section{Conflict of Interest}
The author does not have any conflict of interest to declare.

\bibliographystyle{unsrt}
\bibliography{foundation_models_pathology}

\end{document}